\documentclass[letterpaper]{article}
\usepackage[preprint]{aaai2027}
\usepackage[hyphens]{url}
\usepackage{graphicx}
\usepackage{natbib}
\usepackage{caption}
\usepackage{booktabs}
\usepackage{amsmath}
\usepackage{amssymb}
\usepackage{multirow}
\usepackage{colortbl}

\title{REFLEX: Rethinking MoE Inference as Refinement-Aware Compute Allocation in Diffusion Language Models}

\author{
    Xiang Xia,
    Cheng Yan,
    Yiming Zhang,
    Jiazheng Liu,
    Hongyu Zhang,
    Wuyang Zhang\corresponding
}
\affiliations{
    University of Science and Technology of China, Hefei, China\\
    xxia@mail.ustc.edu.cn, wuyangz@ustc.edu.cn
}

\newcommand{\Kmin}{K_{\min}}
\newcommand{\Kbase}{K_{\mathrm{base}}}
\newcommand{\Khigh}{K_{\max}}
\newcommand{\TopK}{\operatorname{TopK}}
\newcommand{\clip}{\operatorname{clip}}
\newcommand{\mask}{\mathcal{M}}
\newcommand{\score}{S}
\newcommand{\conf}{c}
\newcommand{\wcf}{w_{\mathrm{cf}}}
\newcommand{\vref}{v_{\mathrm{ref}}}
\newcommand{\pairs}{\mathrm{pairs}}
\newcommand{\avgK}{\overline{K}}
\newcommand{\role}{\mathrm{role}}

\begin{document}

\maketitle

\begin{abstract}
Mixture-of-experts (MoE) models increase parameter capacity by activating only a small subset of experts for each token. This conditional-computation paradigm has enabled autoregressive language models to scale model capacity without a proportional increase in per-token computation. In diffusion language models (DLMs), however, each denoising forward jointly revisits all token positions despite their sharply different refinement demands, while the default fixed token-choice routing assigns them a uniform expert budget, creating a mismatch between expert computation and refinement demand. We argue that MoE inference in DLMs should therefore be viewed as refinement-aware compute allocation across heterogeneous token refinement states. We propose REFLEX (\textbf{RE}finement-aware \textbf{FLEX}ible expert allocation), a training-free method that keeps the default router unchanged while reorganizing expert computation around the evolving refinement process. Specifically, REFLEX introduces a coarse-to-fine hierarchy for expert-budget allocation that aligns computation with block-relative refinement roles while using the Frontier-Progress Score to resolve active-block priorities. Across multiple widely used benchmarks on two representative MoE-based DLMs, LLaDA-MoE and LLaDA2.0-mini, REFLEX reduces allocated expert computation by 15\% on average while preserving or even improving generation quality on most benchmarks relative to default routing. Compared with autoregressive-style variable-expert routing methods, REFLEX also yields a more consistent quality--computation trade-off, further supporting the importance of allocating expert computation according to the heterogeneous refinement demands exposed within each denoising forward.

\end{abstract}

\section{Introduction}
\label{sec:introduction}

Mixture-of-experts (MoE) models scale parameter capacity by replacing dense feed-forward layers with multiple experts and routing each token to only a subset of them~\citep{SparselyGatedMoE,GShard,SwitchTransformer}. This conditional-computation design decouples total model capacity from the computation activated for each token. In autoregressive language models~\citep{GPT3}, it has enabled model scaling and strong generation quality without a proportional increase in per-token computation~\citep{GLaM,Mixtral,DeepSeekMoE}.

Diffusion language models (DLMs) have emerged as an alternative generation paradigm that replaces strictly left-to-right prediction with iterative denoising~\citep{D3PM,DiffusionLM,BlockDiffusion}. Rather than generating one token at a time, each denoising forward revisits all token positions and can refine multiple positions in parallel through bidirectional attention. Recent models such as LLaDA~\citep{LLADA}, Dream~\citep{Dream7B}, and SDAR~\citep{SDAR} verify the practical potential of this parallel generation paradigm. To expand model capacity in this setting, MoE architectures have likewise been adopted by scaled DLMs, as exemplified by LLaDA-MoE~\citep{LLaDAMoE} and LLaDA2.0~\citep{LLaDA2}. Consequently, MoE routing now operates over jointly evolving token states rather than a single left-to-right decoding step.

MoE-based DLMs commonly use fixed token-choice (TC) routing, which activates the same number of experts for tokens in different refinement states. However, the shift in generation paradigm creates a new allocation problem for MoE inference: \emph{token states jointly processed within a denoising forward can have substantially different refinement demands, so the same expert budget need not be equally useful for all of them.} Variable-expert strategies developed for autoregressive MoEs determine expert counts from router scores or global budgets~\citep{TopP,SeqTopK,DTopP}. These refinement-agnostic strategies control selected expert counts but do not condition expert-count allocation on token refinement state. In particular, jointly processed states may differ in both their \emph{structural role within iterative diffusion} and their \emph{progress toward commitment}. The resulting mismatch can waste expert computation on states with limited refinement demand while under-allocating it to states where further refinement remains valuable. Recent DLM-specific methods optimize complementary aspects of MoE inference: dMoE~\citep{dMoE} limits unique experts within each block to reduce expert-loading overhead, whereas EC-DLM~\citep{ECDLM} adapts expert capacity across denoising timesteps through a global schedule and expert-choice routing. This leaves open how to allocate a bounded expert budget across heterogeneous refinement states while preserving the default router.

\noindent\textbf{Problem.} How should MoE inference in DLMs allocate expert computation across the heterogeneous token refinement states jointly revisited within each denoising forward?

\noindent\textbf{Contribution.} To address this problem, we rethink MoE inference in DLMs as expert-budget allocation across the token refinement states encountered throughout iterative diffusion. Standard token-choice routing determines which experts process each token, but does not determine how many experts different refinement states should receive. We decouple these decisions: the default router preserves its learned expert ranking, while a refinement-aware allocator assigns expert counts under a per-layer selected-expert budget at each denoising forward. We formalize this decision as a constrained state-wise allocation problem over iterative diffusion, in which heterogeneous refinement states may warrant different amounts of expert computation. \emph{This formulation establishes token refinement state, rather than router scores or denoising progress alone, as the organizing unit of MoE computation allocation.}

To operationalize this formulation, we propose REFLEX (\textbf{RE}finement-aware \textbf{FLEX}ible expert allocation), a training-free method that realizes refinement-aware allocation through a coarse-to-fine hierarchy while leaving the default router unchanged. \emph{REFLEX characterizes token refinement state along two complementary dimensions: its structural role within iterative diffusion and its dynamic progress toward commitment.} Refinement-Role Budgeting converts block-relative roles into a structural allocation prior over all token refinement states processed in a denoising forward. Conditioned on this prior, Frontier-Guided Expert Reallocation uses the Frontier-Progress Score as dynamic evidence to resolve active-block priorities while preserving the block's total expert budget. Across mathematical reasoning and coding benchmarks on LLaDA-MoE and LLaDA2.0-mini, REFLEX reduces selected expert-token pairs by 15\% on average while preserving or even improving generation quality on most benchmarks relative to the default Fixed TC top-8 routing. It also yields a more consistent quality--computation trade-off than autoregressive-style variable-expert routing methods. Together, these results support our central view that MoE inference in DLMs should allocate expert computation according to the heterogeneous refinement demands exposed throughout iterative diffusion.

\section{Related Work}
\label{sec:related_work}

\noindent\textbf{Discrete Diffusion Language Models.}
Discrete diffusion models extend diffusion-based generation to vocabularies through forward corruption and reverse denoising over discrete states~\citep{DDPM,D3PM,MaskGIT}. Diffusion language models (DLMs)~\citep{SEDD,MDLM,ScalingMDM,BlockDiffusion,LLADA} bring iterative denoising to text generation. Unlike the left-to-right prediction of autoregressive models, DLMs use bidirectional attention to refine multiple positions in parallel, offering flexible generation orders and greater parallelism. LLaDA~\citep{LLADA}, Dream~\citep{Dream7B}, and SDAR~\citep{SDAR} have established the viability of this paradigm across language tasks. Recent models advance scale and efficiency. LLaDA2.0~\citep{LLaDA2} scales open DLMs to 100B total parameters, and Seed Diffusion~\citep{SeedDiffusion} achieves high-throughput diffusion decoding for code generation.

\noindent\textbf{MoE in DLMs.}
Sparse MoE layers scale model capacity by routing each token representation to a subset of experts~\citep{SparselyGatedMoE,GShard,SwitchTransformer}. LLaDA-MoE~\citep{LLaDAMoE} and LLaDA2.0~\citep{LLaDA2} extend this conditional architecture to DLMs. Recent work reduces repeated expert activation during denoising. TEAM~\citep{TEAM} tailors activation to decoded, hot, and cold tokens, dMoE~\citep{dMoE} limits unique experts within blocks, and TIDE~\citep{TIDE} schedules expert offloading based on temporal activation stability. EC-DLM~\citep{ECDLM} instead varies expert capacity across denoising timesteps through a global schedule and expert-choice routing. These methods optimize expert reuse, unique-expert activation, offloading, or timestep-level capacity. Per-token expert-count allocation across heterogeneous refinement states within a denoising forward remains underexplored.

\noindent\textbf{Dynamic Routing in Autoregressive MoEs.}
Dynamic routing in autoregressive MoEs varies the number of activated experts according to input difficulty or a computation budget. Threshold-based methods derive expert counts from router scores. Top-$p$~\citep{TopP} uses cumulative routing mass, Expert-Threshold Routing~\citep{ExpertThreshold} selects experts whose routing scores exceed score thresholds, and DTop-$p$~\citep{DTopP} dynamically controls the threshold under a global sparsity constraint. Budget-based methods distribute an activation budget. Alloc-MoE~\citep{AllocMoE} allocates this budget across layers and tokens, while SeqTopK~\citep{SeqTopK} shifts it from individual tokens to the sequence. These methods are tailored to sequential decoding and organize expert allocation around router scores or global budgets. DLM inference revisits token states with heterogeneous refinement roles and progress toward commitment, requiring expert allocation to account for both dimensions.

\section{Preliminaries: Rethinking MoE in DLMs}
\label{sec:prelim}

\subsection{Block-Wise DLM Inference}

Let $\mathbf{u}$ be a prompt with position set $\mathcal{P}$, disjoint from the target positions. A prompt-conditioned DLM generates a target sequence of length $n$ over a vocabulary $\mathcal{V}$ augmented with \texttt{[MASK]}. Block-wise inference partitions the target positions into ordered, non-overlapping blocks $\mathcal{B}=\{\mathcal{B}_1,\ldots,\mathcal{B}_Q\}$ covering $\{1,\ldots,n\}$. We write $b(i)=j$ when target position $i$ belongs to $\mathcal{B}_j$ and set $b(i)=0$ for prompt positions $i\in\mathcal{P}$.

Starting from an all-mask target $\mathbf{x}^{0}$, denoising forward $t$ maintains a partially decoded target $\mathbf{x}^{t}\in(\mathcal{V}\cup\{\texttt{[MASK]}\})^n$, its masked-position set $\mask_t=\{i:x_i^t=\texttt{[MASK]}\}$, and an active-block index $a_t\in\{1,\ldots,Q\}$. The active block is refined until its masked positions are resolved, after which inference advances to the next block.

At masked position $i$, the DLM predicts $p_i^t(\cdot)=p_\phi(\cdot\mid\mathbf{u},\mathbf{x}^t)$, the candidate $\hat{x}_i^t=\operatorname*{arg\,max}_{v\in\mathcal{V}}p_i^t(v)$, and confidence $\conf_i^t=\max_{v\in\mathcal{V}}p_i^t(v)$. Commitment is restricted to masked active-block positions, indexed by $\mathcal{J}_t=\mask_t\cap\mathcal{B}_{a_t}=\{i\in\mask_t:b(i)=a_t\}$.
Let $\mathcal{C}_t\subseteq\mathcal{J}_t$ denote positions committed by the decoding strategy at denoising forward $t$. For the confidence-threshold decoding strategy~\citep{Fast-dLLM}, $\mathcal{C}_t=\{i\in\mathcal{J}_t:\conf_i^t\ge\theta\}$. The target sequence then updates as
\begin{equation}
  x_i^{t+1}=
  \begin{cases}
    \hat{x}_i^t, & i\in\mathcal{C}_t,\\
    \texttt{[MASK]}, & i\in\mathcal{J}_t\setminus\mathcal{C}_t,\\
    x_i^t, & i\notin\mathcal{J}_t.
  \end{cases}
  \label{eq:blockwise_update}
\end{equation}
Blocks are decoded sequentially, while active-block positions are refined and committed in parallel across forwards.

Let $\mathcal{I}_t=\mathcal{P}\cup\{1,\ldots,n\}$ be the set of all prompt and target positions processed by denoising forward $t$. For each $i\in\mathcal{I}_t$, let $s_i^t$ denote the \emph{token refinement state} of position $i$ under the current partial sequence $\mathbf{x}^t$ and, when available, its prediction history. This state describes context-dependent computation during refinement, whether or not the token value remains masked. The same position can induce different states as its context evolves, while states within the same forward can have heterogeneous refinement demands.

\subsection{MoE-Based DLM Inference}

Consider a DLM with MoE feed-forward layers $\mathcal{L}_{\mathrm{moe}}$ and $E$ experts per layer. At layer $\ell\in\mathcal{L}_{\mathrm{moe}}$, state $s_i^t$ has representation $h_{\ell,i}^t$ and router scores $r_{\ell,i,e}^t$ over experts $e\in\{1,\ldots,E\}$. Given an expert count $k_{\ell,i}^t$, token-choice routing selects $\mathcal{T}_{\ell,i}^t=\TopK_e\!\left(r_{\ell,i,e}^t,k_{\ell,i}^t\right)$ and computes
\begin{equation}
  \mathrm{MoE}_{\ell}(h_{\ell,i}^t;k_{\ell,i}^t)
  =
  \sum_{e\in\mathcal{T}_{\ell,i}^t}
  \alpha_{\ell,i,e}^t E_{\ell,e}(h_{\ell,i}^t),
  \label{eq:moe_layer}
\end{equation}
where $E_{\ell,e}(\cdot)$ denotes the $e$-th expert at layer $\ell$, and $\alpha_{\ell,i,e}^t$ is its normalized router weight. Router scores $r_{\ell,i,e}^t$ determine expert preference. The selected expert count $k_{\ell,i}^t$ specifies how many router-ranked experts process token $i$ in state $s_i^t$ at layer $\ell$. Default fixed token-choice routing sets $k_{\ell,i}^t=K$ for every state, whereas variable-count routing allows it to vary.

Per-sample expert computation is measured by selected expert-token pairs, $\pairs=\sum_t\sum_{\ell\in\mathcal{L}_{\mathrm{moe}}}\sum_{i\in\mathcal{I}_t}k_{\ell,i}^t$. Its normalized average is AvgK, $\avgK=\pairs/(|\mathcal{L}_{\mathrm{moe}}|\sum_t|\mathcal{I}_t|)$, the mean selected expert count per token and MoE layer. These metrics quantify selected expert computation, not unique experts loaded into memory.

\subsection{Refinement-Aware MoE Compute Allocation}

We decompose token-choice MoE inference into two decisions: expert ranking and expert-count allocation. Router scores rank experts for each token, while the count specifies how many ranked experts process that token in its refinement state. We preserve router rankings and formulate counts jointly. Let $\mathcal{S}_t=\{s_i^t:i\in\mathcal{I}_t\}$ collect the states processed by forward $t$, and let $\mathbf{k}_t=(k_i^t)_{i\in\mathcal{I}_t}$. Given a per-layer selected-expert budget $K_t^{\Sigma}$, a refinement-aware allocator satisfies
\begin{equation}
  \mathbf{k}_t=\Pi(\mathcal{S}_t,K_t^{\Sigma}),
  \qquad \sum_{i\in\mathcal{I}_t}k_i^t=K_t^{\Sigma},
  \label{eq:allocation_rule}
\end{equation}
where $k_i^t\in\{\Kmin,\ldots,\Khigh\}$. Each count is shared across MoE layers, $k_{\ell,i}^t=k_i^t$, and the default router retains the top $k_i^t$ experts in its score ranking. Default fixed token-choice routing is the uniform special case $k_i^t=K$ and $K_t^{\Sigma}=K|\mathcal{I}_t|$. Thus, the default router still determines which experts are selected, whereas their number is allocated across refinement states under a shared budget.

Let $\mathcal{A}_t$ denote the feasible allocation set defined by Eq.~\eqref{eq:allocation_rule} and the count bounds above. Let $U_i^t(k)$ be the idealized downstream refinement utility of processing token $i$ with $k$ experts in state $s_i^t$. Using an additive state-wise utility surrogate, we write the idealized allocation as
\begin{equation}
  \mathbf{k}_t^\star\in
  \operatorname*{arg\,max}_{\mathbf{k}_t\in\mathcal{A}_t}
  \sum_{i\in\mathcal{I}_t}U_i^t(k_i^t).
  \label{eq:ideal_budget}
\end{equation}
Its $i$-th component, $(\mathbf{k}_t^\star)_i=k_i^{t,\star}$, is the ideal expert count for token $i$ in state $s_i^t$. Define the marginal utility $\Delta U_i^t(k)=U_i^t(k+1)-U_i^t(k)$. Under diminishing returns, the allocation favors tokens with larger marginal utility. Because the budget couples processed tokens, $k_i^{t,\star}$ depends on both the token's refinement state and the refinement demands of other states in $\mathcal{S}_t$, not on router scores alone. Since $U_i^t$ is unobservable, Eq.~\eqref{eq:ideal_budget} defines an allocation principle rather than an executable utility estimator. A practical allocator must therefore organize observable refinement-state information without equating any signal with utility.

\section{Methodology}
\label{sec:method}

To align expert computation with heterogeneous refinement demands during iterative diffusion, we propose REFLEX (\textbf{RE}finement-aware \textbf{FLEX}ible expert allocation), a training-free realization of our formulation that preserves the default router. We overview REFLEX and detail Refinement-Role Budgeting and Frontier-Guided Expert Reallocation.

\subsection{Overview}

\begin{figure*}[t]
\centering
\includegraphics[width=\textwidth]{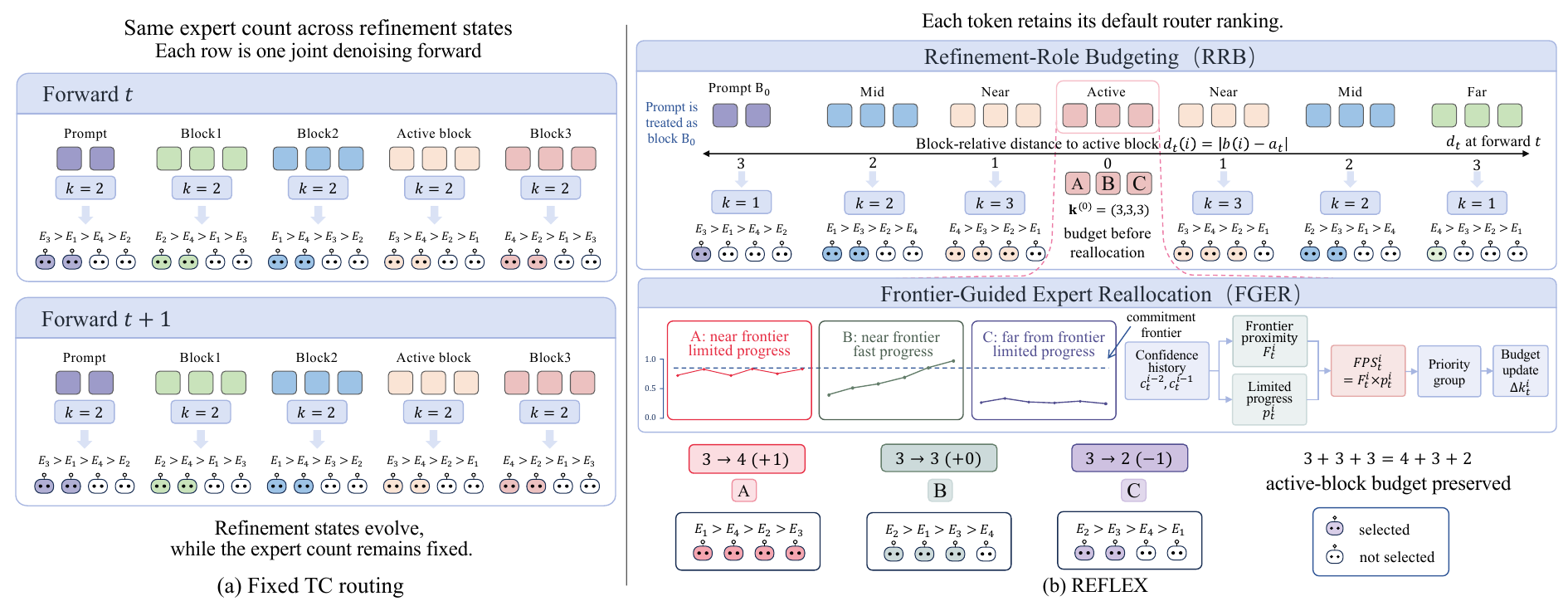}
\caption{Overview of REFLEX. (a) Fixed token-choice routing assigns the same expert count as token refinement states evolve across denoising forwards. (b) REFLEX preserves each token's default router ranking while organizing expert-count allocation through a coarse-to-fine hierarchy. Refinement-Role Budgeting establishes a structural prior from block-relative roles, and Frontier-Guided Expert Reallocation redistributes the active-block budget according to the Frontier-Progress Score to prioritize tokens with greater immediate refinement demand, without changing its total.}
\label{fig:overview}
\end{figure*}

REFLEX instantiates the allocator $\Pi(\mathcal{S}_t,K_t^{\Sigma})$ in Eq.~\eqref{eq:allocation_rule} by organizing observable refinement-state information at two complementary resolutions. Structural role within iterative diffusion establishes an allocation prior, while dynamic progress toward commitment provides evidence to refine it. As illustrated in Figure~\ref{fig:overview}, Refinement-Role Budgeting maps block-relative roles to a structural allocation prior over positions per denoising forward. Given this prior, Frontier-Guided Expert Reallocation uses the Frontier-Progress Score to resolve active-block priorities while preserving the block's expert budget. This coarse-to-fine hierarchy preserves the default router's expert ranking.

\subsection{Refinement-Role Budgeting}

Refinement-Role Budgeting (RRB) establishes the structural allocation prior using the active block as a moving reference for expert computation. States processed in a denoising forward can play different roles relative to that block. RRB therefore organizes expert counts by block-relative role rather than absolute position or forward index.

At denoising forward $t$, the block-relative distance $d_t(i)=|b(i)-a_t|$ describes token $i$'s block-relative role, where $a_t$ is the active-block index and $b(i)$ is the block index of position $i$. We treat prompt positions as block~$0$, enabling prompt tokens to follow the same role-dependent allocation rule as the active block advances during iterative denoising. The base budget $\Kbase$ establishes the nominal level of expert computation, while the role offset $\delta_{\mathrm{role}}$ controls how strongly computation is differentiated across block-relative roles. RRB sets the initial count to $k_{i,\role}^t=H(d_t(i))$, where
\begin{equation}
  H(d)=
    \begin{cases}
      \Kbase+\delta_{\mathrm{role}}, & d\le 1,\\
      \Kbase, & d=2,\\
      \Kbase-\delta_{\mathrm{role}}, & d\ge 3.
    \end{cases}
  \label{eq:role_budget_map}
\end{equation}
For REFLEX, this profile determines the per-layer forward budget $K_t^{\Sigma}=\sum_{i\in\mathcal{I}_t}H(d_t(i))$. The subsequent active-block reallocation preserves this total. The three-level map encodes a block-relative structural allocation prior without treating distance as an estimate of refinement utility. RRB allocates more expert computation to nearby blocks, retains the nominal allocation for intermediate blocks, and reduces it for distant blocks. As $a_t$ advances, each position's computation follows its evolving block-relative role. The resulting profile provides the structural prior for active-block reallocation.

\subsection{Frontier-Guided Expert Reallocation}

The RRB prior accounts for block-relative refinement roles, but unresolved active-block tokens can exhibit different progress toward commitment. Conditioned on this prior, Frontier-Guided Expert Reallocation (FGER) uses confidence history as dynamic evidence for redistributing expert computation within the active block while preserving its total budget. It concentrates computation where commitment proximity coincides with limited confidence progress.

Expert counts for denoising forward $t$ must be determined before computing that forward. FGER therefore infers allocation priorities from confidence history. For each $i\in\mathcal{J}_t$, where $\mathcal{J}_t$ contains unresolved token positions in the active block, it reuses the two latest confidence observations $\conf_i^{t-1}$ and $\conf_i^{t-2}$. The commitment-frontier width $\wcf>0$ controls the frontier interval and maximum extrapolated confidence change. With both observations, the lagged confidence change is $v_i^{t-1}=\conf_i^{t-1}-\conf_i^{t-2}$, clipped to $\bar v_i^{t-1}=\clip(v_i^{t-1},-\wcf/2,\wcf/2)$. Without an additional denoising forward, first-order extrapolation forecasts the unobserved confidence as $\hat{\conf}_i^t=\clip(\conf_i^{t-1}+\bar v_i^{t-1},0,1)$, where $\clip(z,l,u)$ truncates $z$ to $[l,u]$. If $\conf_i^{t-2}$ is unavailable, we set $v_i^{t-1}=0$ and use $\hat{\conf}_i^t=\conf_i^{t-1}$. Clipping prevents an atypical confidence change from dominating the priority estimate.

We next place a smooth gate over the commitment-frontier interval. The commitment threshold $\theta$ is inherited from the decoding strategy. The gate temperature $\tau>0$ controls the smoothness of both commitment proximity and confidence progress, and $\sigma(\cdot)$ denotes the sigmoid function. We define the frontier factor as
\begin{equation}
  F_i^t
  =\sigma\!\left((\hat{\conf}_i^t-\theta+\wcf)/\tau\right)
   \sigma\!\left((\theta-\hat{\conf}_i^t)/\tau\right).
  \label{eq:frontier_factor}
\end{equation}
The frontier factor $F_i^t$ assigns high values for forecasts within the commitment-frontier interval $[\theta-\wcf,\theta]$, emphasizing tokens whose refinement can affect the commitment decision. To distinguish tokens resolving from those making limited progress, the velocity reference $\vref$ sets the boundary below which confidence progress is treated as slow. We define
\begin{equation}
  P_i^t
  =\sigma\!\left((\vref-\bar v_i^{t-1})/\tau\right).
  \label{eq:progress_factor}
\end{equation}
The factor $P_i^t$ encodes limited confidence progress, increasing as $\bar v_i^{t-1}$ falls below $\vref$. These conditions jointly define the Frontier-Progress Score (FPS):
\begin{equation}
  \score_i^t=F_i^tP_i^t.
  \label{eq:frontier_progress_score}
\end{equation}
The multiplicative form suppresses tokens satisfying only one condition, so proximity or slow progress alone is insufficient evidence of refinement demand. When $\conf_i^{t-2}$ is unavailable, the forecast above uses zero velocity and we set $P_i^t=1$, yielding frontier-only prioritization. We use $\score_i^t$ to rank active-block tokens from dynamic evidence, not to estimate marginal utility in Eq.~\eqref{eq:ideal_budget}.

Finally, FGER converts this priority ordering into expert counts while preserving the total active-block expert budget established by RRB. The quantity $K_{\mathrm{cur}}=\Kbase+\delta_{\mathrm{role}}$ is the RRB count for an active-block token. The reallocation offset $\delta_{\mathrm{cur}}$ controls the strength of computation redistribution, while $q\in(0,1/2]$ controls its coverage among unresolved tokens. Ranking $\mathcal{J}_t$ by $\score_i^t$ partitions it into high-, middle-, and low-priority groups $\mathcal{G}_{\mathrm{high}}$, $\mathcal{G}_{\mathrm{mid}}$, and $\mathcal{G}_{\mathrm{low}}$ with proportions $q$, $1-2q$, and $q$. FGER assigns
\begin{equation}
  k_i^t=
  \begin{cases}
    K_{\mathrm{cur}}+\delta_{\mathrm{cur}}, & i\in\mathcal{G}_{\mathrm{high}},\\
    K_{\mathrm{cur}}, & i\in\mathcal{G}_{\mathrm{mid}},\\
    K_{\mathrm{cur}}-\delta_{\mathrm{cur}}, & i\in\mathcal{G}_{\mathrm{low}},
  \end{cases}
  \label{eq:current_reallocation}
\end{equation}
where $k_i^t$ is the expert count for token $i$ at forward $t$. Without confidence history, active-block tokens retain $K_{\mathrm{cur}}$; other tokens retain the RRB count in Eq.~\eqref{eq:role_budget_map}. Equal-sized high- and low-priority groups have canceling count changes, so $\sum_{i\in\mathcal{J}_t}k_i^t=|\mathcal{J}_t|K_{\mathrm{cur}}$. FGER redistributes active-block computation without changing its total budget. This coarse-to-fine allocation combines a block-relative structural allocation prior with dynamic evidence from progress toward commitment while preserving the default router's expert ranking.

\section{Experiments}
\label{sec:experiments}

We evaluate whether REFLEX improves the trade-off between generation quality and allocated expert computation on two representative MoE-based DLMs across multiple benchmarks. This section presents the experimental setup, comparative results, ablation studies, and hyper-parameter analysis. The experiments address four main questions:

\begin{enumerate}
  \item[\textbf{Q1:}] Can REFLEX preserve or even improve quality with less expert computation than Fixed TC routing?
  \item[\textbf{Q2:}] Is REFLEX better suited to MoE-based DLM inference than existing variable-expert allocation strategies?
  \item[\textbf{Q3:}] How does modeling each refinement-state dimension influence the quality--computation trade-off of REFLEX?
  \item[\textbf{Q4:}] How do hyper-parameter changes affect the expert allocation behavior of REFLEX?
\end{enumerate}

\newcommand{\accgain}[1]{$_{{\color{green!45!black}\scriptstyle +#1}}$}
\newcommand{\accloss}[1]{$_{{\color{red!70!black}\scriptstyle -#1}}$}
\newcommand{\accsame}[1]{$_{{\color{gray}\scriptstyle #1}}$}
\newcommand{\pairred}[1]{$_{{\color{green!45!black}\scriptstyle \downarrow #1\%}}$}
\newcommand{\pairinc}[1]{$_{{\color{red!70!black}\scriptstyle \uparrow #1\%}}$}

\subsection{Experimental Setup}

\noindent\textbf{Models, tasks, and metrics.}
We evaluate two representative publicly available MoE-based DLMs, LLaDA-MoE (7B)~\citep{LLaDAMoE} and LLaDA2.0-mini (16B)~\citep{LLaDA2}, on GSM8K 5-shot~\citep{GSM8K}, Math500 4-shot~\citep{Math500}, HumanEval 0-shot~\citep{HumanEval}, and MBPP 3-shot~\citep{MBPP}. We use task accuracy or pass@1 to evaluate generation quality, and use AvgK, the average selected expert count per token at each MoE layer, together with selected expert-token pairs per sample to evaluate expert computation. For a fixed model and processed sequence length, $\mathrm{Pairs}\propto\mathrm{AvgK}\times\mathrm{NFE}$, where NFE is the number of model forward passes per sample. Appendix~A explains how selected expert-token pairs reflect routed-expert linear-layer FLOPs.

\noindent\textbf{Compared methods.}
We first compare REFLEX with three types of token-choice strategies: fixed, budget-based, and threshold-based. Fixed TC selects a fixed number of token-choice experts per token. Budgeted TC~\citep{AllocMoE,SeqTopK} assigns a variable number of token-choice experts under a target average expert budget. Threshold-based strategies select experts according to routing-score thresholds: Top-$p$~\citep{TopP} selects experts until the cumulative routing mass reaches a threshold, DTop-$p$~\citep{DTopP} adjusts this threshold to target an average budget, and Expert-Threshold Routing~\citep{ExpertThreshold} selects experts whose routing scores exceed a global threshold. These methods vary expert count without modeling diffusion refinement roles. Because EC-DLM~\citep{ECDLM} replaces token-choice routing with expert-choice routing, we further compare REFLEX with an EC-DLM-style late-high timestep schedule under the default router.

\noindent\textbf{Experimental setting.}
For a fair comparison, all experiments use the following hyper-parameter settings unless otherwise specified. All experiments are conducted on NVIDIA A100 80GB GPUs. All methods are evaluated with the lm-evaluation-harness under deterministic decoding with temperature 0. All methods use generation length $n=256$, block length $n_{\mathrm{blk}}=32$, and commitment threshold $\theta=0.9$. For REFLEX, Refinement-Role Budgeting (RRB) uses base role budget $K_{\mathrm{base}}=8$ and role offset $\delta_{\mathrm{role}}=4$. Frontier-Guided Expert Reallocation (FGER) uses active-block reallocation offset $\delta_{\mathrm{cur}}=2$ and outer-group fraction $q=0.25$. The Frontier-Progress Score (FPS) uses velocity reference $v_{\mathrm{ref}}=0.01$ and gate temperature $\tau=0.05$. The commitment-frontier width is $\wcf=0.30$ for LLaDA-MoE and $\wcf=0.25$ for LLaDA2.0-mini. For comparison strategies, Fixed TC includes top-4 and top-8. Budgeted TC uses target AvgK 6 and 7 for its low/high settings, respectively. Threshold-based strategies are calibrated to low/high computation levels around AvgK 6 and 8. Appendix~B details the comparison strategies and implementation settings.

\subsection{Main Results}

\noindent\textbf{To answer Q1.} Table~\ref{tab:main_results} compares REFLEX with default Fixed TC top-8 and its lower-computation top-4 setting. Top-8 tests whether REFLEX can preserve or even improve default routing quality with less expert computation, while top-4 provides a fixed-count reference at a lower budget.

\begin{table*}[t]
  \centering
  {
  \small
  \setlength{\tabcolsep}{1mm}
    \begin{tabular}{ccccccccccc}
    \toprule
    \multirow{2}{*}{\textbf{Family}} &
    \multirow{2}{*}{\textbf{Method}} &
    \multicolumn{2}{c}{\textbf{GSM8K (5-shot)}} &
    \multicolumn{2}{c}{\textbf{Math500 (4-shot)}} &
    \multicolumn{2}{c}{\textbf{MBPP (3-shot)}} &
    \multicolumn{2}{c}{\textbf{HumanEval (0-shot)}} &
    \multirow{2}{*}{\textbf{AvgK}} \\
    \cmidrule(lr){3-4}\cmidrule(lr){5-6}\cmidrule(lr){7-8}\cmidrule(lr){9-10}
    & & Acc $\Uparrow$ & Pairs (M) $\Downarrow$
      & Acc $\Uparrow$ & Pairs (M) $\Downarrow$
      & Acc $\Uparrow$ & Pairs (M) $\Downarrow$
      & Acc $\Uparrow$ & Pairs (M) $\Downarrow$ & \\
    \midrule
    \rowcolor{lightgray}
    \multicolumn{11}{c}{\textbf{LLaDA-MoE}} \\
    \midrule
    \multirow{2}{*}{Fixed}
      & Fixed TC (top-4) & 66.8\accloss{9.7} & 16.3\pairred{49} & 34.6\accloss{4.6} & 12.6\pairred{50} & 44.8\accloss{10.0} & 8.9\pairred{48} & 43.9\accloss{8.5} & 5.0\pairred{49} & 4.0 \\
      & Fixed TC (top-8) & 76.5 & 32.2 & 39.2 & 25.3 & 54.8 & 17.2 & 52.4 & 9.8 & 8.0 \\
    \midrule
    \multirow{2}{*}{Budget}
      & Budgeted TC (low) & 76.8\accgain{0.3} & 24.6\pairred{23} & 38.0\accloss{1.2} & 19.6\pairred{22} & 51.4\accloss{3.4} & 13.4\pairred{22} & 53.7\accgain{1.3} & 7.1\pairred{27} & 6.0 \\
      & Budgeted TC (high) & 76.2\accloss{0.3} & 28.7\pairred{11} & 38.6\accloss{0.6} & 22.9\pairred{9} & 52.0\accloss{2.8} & 13.5\pairred{22} & 49.4\accloss{3.0} & 8.2\pairred{16} & 7.0 \\
    \midrule
    \multirow{6}{*}{Threshold}
      & Top-$p$ (low) & 67.3\accloss{9.2} & 26.5\pairred{18} & 31.6\accloss{7.6} & 19.5\pairred{23} & 40.0\accloss{14.8} & 13.1\pairred{24} & 33.5\accloss{18.9} & 7.1\pairred{28} & 6.0 \\
      & Top-$p$ (high) & 73.5\accloss{3.0} & 34.8\pairinc{8} & 39.0\accloss{0.2} & 26.5\pairinc{5} & 49.2\accloss{5.6} & 16.4\pairred{5} & 52.4\accsame{0.0} & 9.3\pairred{6} & 8.1 \\
      & DTop-$p$ (low) & 71.2\accloss{5.3} & 24.1\pairred{25} & 36.4\accloss{2.8} & 18.7\pairred{26} & 43.6\accloss{11.2} & 14.3\pairred{17} & 42.7\accloss{9.7} & 6.9\pairred{29} & 6.0 \\
      & DTop-$p$ (high) & 75.4\accloss{1.1} & 31.2\pairred{3} & 38.8\accloss{0.4} & 25.0\pairred{1} & 52.2\accloss{2.6} & 16.5\pairred{4} & 52.4\accsame{0.0} & 9.2\pairred{6} & 8.0 \\
      & Expert-thr (low) & 70.7\accloss{5.8} & 24.4\pairred{24} & 37.2\accloss{2.0} & 20.5\pairred{19} & 46.8\accloss{8.0} & 8.4\pairred{52} & 43.3\accloss{9.1} & 8.1\pairred{18} & 6.0 \\
      & Expert-thr (high) & 76.5\accsame{0.0} & 33.5\pairinc{4} & 40.8\accgain{1.6} & 27.3\pairinc{8} & 51.4\accloss{3.4} & 15.9\pairred{8} & 49.4\accloss{3.0} & 10.3\pairinc{5} & 8.1 \\
    \midrule
    \multicolumn{2}{c}{EC-DLM-style}
      & \textbf{77.0}\accgain{0.5} & 32.7\pairinc{1} & 40.0\accgain{0.8} & 26.2\pairinc{4} & 50.4\accloss{4.4} & 19.4\pairinc{13} & 51.8\accloss{0.6} & 10.3\pairinc{5} & 8.2 \\
    \midrule
    \rowcolor{blue!5}
    \multicolumn{2}{c}{\textbf{REFLEX}}
      & 76.4\accloss{0.1} & 27.7\pairred{14} & \textbf{41.8}\accgain{2.6} & 21.5\pairred{15} & \textbf{55.2}\accgain{0.4} & 12.9\pairred{25} & \textbf{54.3}\accgain{1.9} & 9.0\pairred{9} & 6.6 \\
    \midrule
    \rowcolor{lightgray}
    \multicolumn{11}{c}{\textbf{LLaDA2.0-mini}} \\
    \midrule
    \multirow{2}{*}{Fixed}
      & Fixed TC (top-4) & 61.0\accloss{22.8} & 17.5\pairred{32} & 20.0\accloss{3.0} & 12.3\pairred{46} & 45.0\accloss{10.0} & 10.0\pairred{23} & 15.2\accloss{19.6} & 6.0\pairred{37} & 4.0 \\
      & Fixed TC (top-8) & 83.8 & 25.8 & 23.0 & 22.7 & 55.0 & 12.9 & 34.8 & 9.6 & 8.0 \\
    \midrule
    \multirow{2}{*}{Budget}
      & Budgeted TC (low) & 82.6\accloss{1.2} & 20.8\pairred{19} & 23.8\accgain{0.8} & 17.3\pairred{24} & 53.4\accloss{1.6} & 11.4\pairred{12} & 28.0\accloss{6.8} & 7.6\pairred{20} & 6.0 \\
      & Budgeted TC (high) & 82.9\accloss{0.9} & 23.0\pairred{11} & 22.8\accloss{0.2} & 19.8\pairred{13} & 54.8\accloss{0.2} & 11.6\pairred{10} & 32.9\accloss{1.9} & 9.0\pairred{6} & 7.0 \\
    \midrule
    \multirow{6}{*}{Threshold}
      & Top-$p$ (low) & 70.3\accloss{13.5} & 24.2\pairred{6} & 19.6\accloss{3.4} & 17.3\pairred{24} & 44.0\accloss{11.0} & 12.0\pairred{7} & 15.2\accloss{19.6} & 9.4\pairred{2} & 6.3 \\
      & Top-$p$ (high) & 81.4\accloss{2.4} & 27.6\pairinc{7} & 24.2\accgain{1.2} & 21.5\pairred{5} & 52.6\accloss{2.4} & 13.4\pairinc{4} & 25.0\accloss{9.8} & 10.7\pairinc{12} & 8.1 \\
      & DTop-$p$ (low) & 64.9\accloss{18.9} & 25.1\pairred{3} & 19.6\accloss{3.4} & 17.9\pairred{21} & 48.8\accloss{6.2} & 13.2\pairinc{2} & 15.9\accloss{18.9} & 9.9\pairinc{4} & 6.0 \\
      & DTop-$p$ (high) & 78.5\accloss{5.3} & 30.1\pairinc{16} & 23.6\accgain{0.6} & 22.6\pairred{1} & 55.0\accsame{0.0} & 14.2\pairinc{10} & 20.1\accloss{14.7} & 11.3\pairinc{19} & 8.0 \\
      & Expert-thr (low) & 81.7\accloss{2.1} & 21.7\pairred{16} & 23.8\accgain{0.8} & 18.3\pairred{19} & 55.8\accgain{0.8} & 12.7\pairred{2} & 28.0\accloss{6.8} & 7.9\pairred{17} & 6.2 \\
      & Expert-thr (high) & \textbf{84.9}\accgain{1.1} & 26.4\pairinc{2} & 23.2\accgain{0.2} & 23.0\pairinc{1} & 54.6\accloss{0.4} & 13.9\pairinc{8} & 33.5\accloss{1.3} & 10.1\pairinc{6} & 8.0 \\
    \midrule
    \multicolumn{2}{c}{EC-DLM-style}
      & 74.9\accloss{8.9} & 28.5\pairinc{10} & 22.0\accloss{1.0} & 23.6\pairinc{4} & 52.8\accloss{2.2} & 15.3\pairinc{19} & 23.2\accloss{11.6} & 10.5\pairinc{9} & 7.9 \\
    \midrule
    \rowcolor{blue!5}
    \multicolumn{2}{c}{\textbf{REFLEX}}
      & 82.3\accloss{1.5} & 23.2\pairred{10} & \textbf{24.8}\accgain{1.8} & 18.2\pairred{20} & \textbf{56.2}\accgain{1.2} & 10.3\pairred{21} & \textbf{38.4}\accgain{3.6} & 8.9\pairred{7} & 6.5 \\
    \bottomrule
    \end{tabular}
  }
  \caption{Main results on LLaDA-MoE and LLaDA2.0-mini. For each task, we report accuracy (Acc, \%), using pass@1 for coding tasks, and selected expert-token pairs per sample in millions; AvgK is averaged across tasks. Expert-thr denotes Expert-Threshold Routing. Acc and Pairs subscripts denote percentage-point and relative changes from Fixed TC (top-8), respectively; \textcolor{green!45!black}{green} indicates higher Acc or fewer Pairs, and \textcolor{red!70!black}{red} indicates the reverse. Best accuracies are shown in bold.}
  \label{tab:main_results}
\end{table*}

Relative to default Fixed TC top-8, REFLEX consistently reduces selected expert-token pairs. On most benchmarks, REFLEX improves generation quality despite the lower budget, while using 15\% fewer pairs on average across all settings. The main boundary case is LLaDA2.0-mini GSM8K, where REFLEX uses 10\% fewer pairs but loses 1.5 points. Although AvgK decreases by 19.5\%, NFE increases by 11.6\%, offsetting part of the expected computation reduction without fully preserving quality. By contrast, Fixed TC top-4 also reduces expert computation but degrades every task on both models. Both reduce expert computation, but only REFLEX conditions this reduction on token refinement state. \emph{A lower uniform budget therefore cannot explain the quality--computation improvement of REFLEX.}

This reflects the coarse-to-fine allocation of REFLEX. Refinement-Role Budgeting varies expert counts with block-relative structural role, while Frontier-Guided Expert Reallocation redistributes the fixed active-block budget according to progress toward commitment. REFLEX therefore reduces overall computation without imposing the same reduction on all refinement states. Appendix~A further analyzes how the resulting allocation determines the computation reduction. In a nutshell, REFLEX improves the Fixed TC quality--computation trade-off by aligning expert computation with heterogeneous refinement demands, which answers Q1.

\begin{table*}[t]
\centering
{
\small
\setlength{\tabcolsep}{1mm}
\begin{tabular}{cccccccccc}
\toprule
\multirow{2}{*}{\textbf{Configuration}} &
\multicolumn{2}{c}{\textbf{GSM8K (5-shot)}} &
\multicolumn{2}{c}{\textbf{Math500 (4-shot)}} &
\multicolumn{2}{c}{\textbf{MBPP (3-shot)}} &
\multicolumn{2}{c}{\textbf{HumanEval (0-shot)}} &
\multirow{2}{*}{\textbf{AvgK}} \\
\cmidrule(lr){2-3}\cmidrule(lr){4-5}\cmidrule(lr){6-7}\cmidrule(lr){8-9}
& Acc $\Uparrow$ & Pairs (M) $\Downarrow$
& Acc $\Uparrow$ & Pairs (M) $\Downarrow$
& Acc $\Uparrow$ & Pairs (M) $\Downarrow$
& Acc $\Uparrow$ & Pairs (M) $\Downarrow$ & \\
\midrule
Fixed TC (top-8)
  & 76.5 & 32.2 & 39.2 & 25.3 & 54.8 & 17.2 & 52.4 & 9.8 & 8.0 \\
RRB Only
  & 76.0\accloss{0.5} & 27.6\pairred{14}
  & 40.2\accgain{1.0} & 21.5\pairred{15}
  & 53.6\accloss{1.2} & 12.9\pairred{25}
  & \textbf{55.5}\accgain{3.1} & 8.8\pairred{10} & 6.6 \\
FGER Only
  & \textbf{77.9}\accgain{1.4} & 32.6\pairinc{1}
  & 40.2\accgain{1.0} & 25.8\pairinc{2}
  & 54.0\accloss{0.8} & 17.5\pairinc{2}
  & 54.3\accgain{1.9} & 9.6\pairred{2} & 8.0 \\
\rowcolor{blue!6}
\textbf{REFLEX}
  & 76.4\accloss{0.1} & 27.7\pairred{14}
  & \textbf{41.8}\accgain{2.6} & 21.5\pairred{15}
  & \textbf{55.2}\accgain{0.4} & 12.9\pairred{25}
  & 54.3\accgain{1.9} & 9.0\pairred{9} & 6.6 \\
\bottomrule
\end{tabular}
}
\caption{Component ablation on LLaDA-MoE. For each task, we report accuracy, using pass@1 for coding tasks, and selected expert-token pairs per sample in millions; AvgK is averaged across tasks. Accuracy subscripts are percentage-point changes from Fixed TC (top-8), while Pairs subscripts are relative changes. All other settings are identical to those in Table~\ref{tab:main_results}.}
\label{tab:component_ablation}
\end{table*}

\noindent\textbf{To answer Q2.} As shown in Table~\ref{tab:main_results}, we compare REFLEX with autoregressive-style variable-expert methods and a DLM-specific alternative. Budgeted TC varies expert counts under a global budget, whereas Top-$p$, DTop-$p$, and Expert-Threshold Routing derive them from router-score statistics. The EC-DLM-style late-high schedule varies expert counts with denoising progress under the default router. Together, these strategies test whether variable-expert criteria effectively allocate computation for MoE-based DLM inference.

\begin{figure}[t]
  \centering
  \includegraphics[width=\columnwidth]{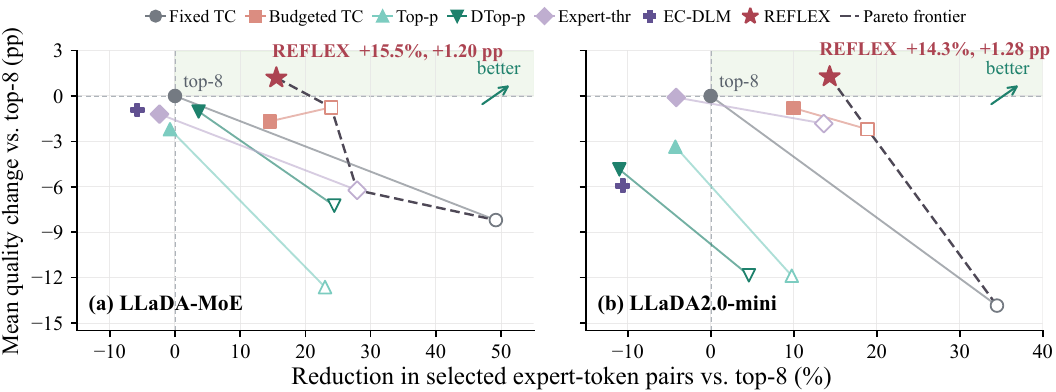}
  \caption{Cross-task quality--computation trade-off on LLaDA-MoE and LLaDA2.0-mini. Each point averages four tasks, and both axes report changes from Fixed TC top-8. Hollow and filled markers denote low and high settings, respectively; dashed lines show the empirical Pareto frontier.}
  \label{fig:pareto_tradeoff}
\end{figure}

Figure~\ref{fig:pareto_tradeoff} summarizes the cross-task quality--computation trade-off. Across the four tasks, REFLEX improves the aggregate trade-off over default Fixed TC top-8 on both models, reducing selected expert-token pairs by 15.5\% and 14.3\%, respectively, while also improving average generation quality. Its results also lie on the empirical Pareto frontier among the evaluated strategies. Competing strategies obtain larger computation reductions only with lower average quality, or approach default quality while forfeiting most of the computation saving. The EC-DLM-style schedule similarly provides a weaker trade-off and increases selected pairs on both models. \emph{The advantage of REFLEX therefore cannot be attributed to varying expert counts or adapting them to global denoising progress alone.}

These differences reflect how each method allocates expert computation. Budgeted TC controls expert computation without conditioning it on refinement state. Threshold-based methods derive expert counts from router scores, which express expert preference rather than refinement demand. The EC-DLM-style schedule adapts expert counts to denoising progress, but not to refinement-state differences within the forward. REFLEX conditions expert-count allocation on structural role within iterative diffusion and progress toward commitment while preserving the default router's ranking. In a nutshell, REFLEX is better suited to MoE-based DLM inference because it explicitly aligns expert allocation with heterogeneous token refinement demands, which answers Q2.

\subsection{Ablation of Allocation Components}

\noindent\textbf{To answer Q3.} Table~\ref{tab:component_ablation} separates the two allocation levels of REFLEX. RRB Only applies the block-relative profile without active-block reallocation, whereas FGER Only reallocates the active-block budget over uniform top-8 allocation.

RRB Only reduces selected pairs by 10--25\% across all tasks while improving Math500 and HumanEval. This shows that block-relative refinement role identifies where expert counts can be reduced without imposing a uniform reduction. FGER Only improves three tasks while retaining AvgK 8 and essentially the same selected-pair count as Fixed TC top-8. It therefore improves where active-block computation is placed, but does not itself reduce the overall budget. Combining both levels improves three tasks over RRB Only at the same AvgK and nearly identical selected-pair counts. \emph{RRB accounts for the computation reduction, whereas FGER improves allocation within the resulting structural profile.}

Appendix~C further validates RRB's block-relative profile and compares FGER ranking signals at matched computation. FPS leads on both tasks, supporting joint use of commitment proximity and confidence progress. In a nutshell, RRB establishes the computation-saving profile, while FPS improves active-block allocation, which answers Q3.

\subsection{Hyper-parameter Analysis}

\noindent\textbf{To answer Q4.} To examine the expert allocation behavior of REFLEX across hyper-parameter settings, we vary the commitment-frontier width $\wcf$, velocity reference $\vref$, gate temperature $\tau$, and outer-group fraction $q$ on LLaDA-MoE. Detailed results and analysis are provided in Appendix~D. In a nutshell, REFLEX shows broadly consistent allocation behavior across the tested hyper-parameter values, while quality remains task dependent, which answers Q4.

\section{Conclusion and Discussion}
\label{sec:conclusion_discussion}

\noindent\textbf{Conclusion.}
We rethink MoE inference in DLMs as refinement-aware expert-budget allocation across token refinement states during iterative diffusion. REFLEX realizes this view through two complementary dimensions. Refinement-Role Budgeting organizes expert computation by block-relative structural role, while Frontier-Guided Expert Reallocation resolves active-block priorities from progress toward commitment through the Frontier-Progress Score. Across two representative MoE-based DLMs, REFLEX achieves a better quality--computation trade-off than default Fixed TC routing and autoregressive-style variable-expert methods. These results support allocating MoE computation according to refinement state rather than applying a uniform expert budget throughout diffusion generation.

\noindent\textbf{Discussion.}
REFLEX performs refinement-aware expert-budget allocation across token refinement states. Accordingly, reductions in selected expert-token pairs quantify allocated expert computation rather than end-to-end latency, which depends on unique-expert activation, memory traffic, and kernel efficiency. These complementary dimensions suggest that REFLEX could be combined with dMoE~\citep{dMoE} and fused expert dispatch. The current formulation shares the expert count $k_i^t$ across all MoE layers. A layer-dependent allocation $k_{\ell,i}^t$ could account for depth-specific expert specialization, but would require estimating refinement demand jointly across tokens and layers under a shared expert budget. EC-DLM~\citep{ECDLM} explores expert-choice routing for DLM inference. Extending refinement-aware expert-budget allocation to this routing paradigm is therefore a promising direction.

\bibliography{references}

\clearpage
\setcounter{secnumdepth}{1}
\appendix
\renewcommand{\thetable}{\Alph{table}}
\setcounter{table}{0}
This appendix provides supporting analyses and additional experimental evidence for REFLEX. Appendix~A connects selected expert-token pairs to routed-expert linear-layer FLOPs and decomposes the realized computation reduction. Appendix~B details the comparison strategies and their settings. Appendix~C examines block-relative allocation and active-block reallocation signals. Appendix~D analyzes the behavior of REFLEX under different hyper-parameter settings.

\section{Analysis of Allocated Expert Computation}

\noindent\textbf{Selected expert-token pairs and routed-expert FLOPs.}
For a SwiGLU routed expert with hidden size $h$ and expert intermediate width $d_{\mathrm{exp}}$, each selected expert-token pair corresponds to two input projections and one output projection. Counting a multiplication and addition as two FLOPs, the routed-expert linear-layer cost is
\begin{equation}
  C_{\mathrm{routed}}^{\mathrm{linear}}
  = 6h d_{\mathrm{exp}}\,\pairs.
  \label{eq:routed_expert_flops}
\end{equation}
Within a fixed model, $h$ and $d_{\mathrm{exp}}$ are constant, so Eq.~\eqref{eq:routed_expert_flops} makes relative changes in selected expert-token pairs identical to those in routed-expert linear-layer FLOPs. The normalized metric AvgK separates the selected expert count from the number of processed states. Letting $\mathcal{L}_{\mathrm{moe}}$ denote the MoE layers, its definition gives $\pairs=|\mathcal{L}_{\mathrm{moe}}|\avgK\sum_t|\mathcal{I}_t|$. Thus, realized selected pairs depend on both the average expert count assigned to a processed token and the number of token refinement states visited along its generation trajectory. For fixed prompt and generation lengths, the latter is governed primarily by NFE, the number of denoising forwards. Selected pairs exclude attention, router scoring, shared experts, and dispatch overhead, and therefore measure selected routed-expert linear-layer computation rather than end-to-end FLOPs or latency.

\noindent\textbf{Sources of computation reduction.}
Refinement-Role Budgeting (RRB) establishes the expert-count profile across block-relative roles and directly determines the reduction in AvgK. Frontier-Guided Expert Reallocation (FGER) then assigns equal-sized high- and low-priority groups around the active-block count established by RRB, so the positive and negative count changes cancel within that block. FGER consequently changes where active-block computation is assigned without changing its total budget for the current denoising forward. Its allocation can nevertheless influence which tokens are committed and hence the number of states processed in later forwards. The final reduction in selected expert-token pairs therefore combines a direct per-state effect from the lower RRB profile with an indirect trajectory effect through NFE. This decomposition explains why configurations with similar AvgK can realize slightly different pair counts and why AvgK alone does not completely determine per-sample computation.

\section{Compared Methods and Settings}

All comparison strategies use the same model checkpoints, evaluation prompts, decoding strategy, and default router ranking as REFLEX. They differ in how the number of router-ranked experts is assigned to each token. This controlled setup isolates the expert-count allocation criterion while retaining the model's learned expert preferences.

\noindent\textbf{Fixed TC.}
Fixed TC applies the default token-choice router with the same selected expert count for every token, regardless of refinement state. We report top-4 as a lower-computation uniform reference and top-8 as the default routing setting. Their comparison distinguishes refinement-aware allocation from simply lowering the expert count everywhere.

\noindent\textbf{Budgeted TC.}
Budgeted TC is one comparison method implemented with a shared budget-allocation rule for both configurations~\citep{AllocMoE,SeqTopK}. It distributes variable expert counts under a sequence-level average budget using router-derived priorities. The low and high configurations differ only in their target average budgets, with target AvgK 6 and 7, respectively. Budgeted TC therefore tests whether controlling total expert computation is sufficient without explicitly conditioning the allocation on structural role within iterative diffusion or progress toward commitment.

\noindent\textbf{Threshold-based methods.}
Threshold-based methods derive selected expert counts directly from router-score statistics. To keep their computation range comparable, all threshold-based configurations use selected expert-count bounds $[1,12]$.
\begin{itemize}
  \item \textbf{Top-$p$}~\citep{TopP} selects the smallest prefix of router-ranked experts whose cumulative routing mass reaches a threshold. Its low/high thresholds are $p=0.15/0.20$ for LLaDA-MoE and $p=0.25/0.325$ for LLaDA2.0-mini.
  \item \textbf{DTop-$p$}~\citep{DTopP} adapts the cumulative-mass threshold online to target AvgK 6 or 8. We constrain the threshold to $[0.01,0.95]$.
  \item \textbf{Expert-Threshold Routing}~\citep{ExpertThreshold}, abbreviated as Expert-thr in the main table, retains experts whose individual router weights exceed a calibrated global threshold. Its low/high thresholds are $0.0215/0.0190$ for LLaDA-MoE and $0.0270/0.0220$ for LLaDA2.0-mini.
\end{itemize}
Although their thresholding rules differ, all three strategies treat router scores as the criterion for varying expert counts. They therefore evaluate whether expert preference alone provides an adequate signal for expert-count allocation during iterative denoising.

\noindent\textbf{EC-DLM-style timestep allocation.}
EC-DLM~\citep{ECDLM} combines expert-choice routing with timestep-dependent capacity and allocates more capacity at low-mask-ratio timesteps. To isolate its temporal allocation criterion without changing the routing type, we implement an EC-DLM-style late-high schedule under the default token-choice router. The schedule increases the selected expert count from 4 to 12 as denoising progresses, assigning more computation to later forwards at the global schedule level. This comparison is not a reproduction of EC-DLM because it retains token-choice routing and requires no retraining; it tests whether global denoising progress alone provides a sufficient allocation criterion under the routing interface shared by the other methods.

\noindent\textbf{Scope relative to block-level expert aggregation.}
REFLEX is complementary to methods such as dMoE~\citep{dMoE}, which reduce unique-expert activation within a decoding block to lower expert-loading overhead. Such methods optimize which distinct experts must be loaded or dispatched, whereas REFLEX controls how many router-ranked experts are assigned across token refinement states. We therefore report selected expert-token pairs, which measure the allocation directly controlled by REFLEX, rather than unique-expert counts or memory traffic.

\section{Analysis of Refinement-Aware Allocation}

\noindent\textbf{Block-relative allocation behavior.}
RRB uses the active block as a moving reference and treats prompt positions as block~0. It assigns more expert computation to tokens in the active and nearby blocks, retains the nominal count at intermediate block-relative roles, and reduces computation for distant blocks. As the active block advances, a position's allocation changes with its evolving structural role rather than its absolute index. This profile is a structural allocation prior, not an estimate of latent refinement utility. Conditioned on that prior, FGER uses dynamic evidence within the active block to redistribute expert counts across priority groups while preserving the total RRB budget.

\noindent\textbf{Validation of block-relative allocation.}
To test whether the block-relative correspondence of RRB matters beyond its expert-count distribution, we construct Role-Shuffled RRB from RRB Only. At every denoising forward, this intervention preserves the multiset of selected expert counts and their total budget but randomly reassigns those counts among the processed token positions. It therefore removes the correspondence between expert count and block-relative structural role without changing the available computation for that forward. We vary only the allocation seed and report results over three seeds.

\begin{table}[t]
\centering
{
\small
\setlength{\tabcolsep}{1mm}
\begin{tabular}{ccccc}
\toprule
\textbf{Benchmark} & \textbf{Configuration} & \textbf{Acc} & \textbf{AvgK} & \textbf{Pairs (M)} \\
\midrule
\multirow{2}{*}{\textbf{Math500}}
  & \cellcolor{blue!6}\textbf{RRB Only}
  & \cellcolor{blue!6}\textbf{40.2}
  & \cellcolor{blue!6}6.28
  & \cellcolor{blue!6}21.5 \\
  & Role-Shuffled RRB & $36.5{\pm}1.9$ & 6.29 & 20.0 \\
\midrule
\multirow{2}{*}{\textbf{HumanEval}}
  & \cellcolor{blue!6}\textbf{RRB Only}
  & \cellcolor{blue!6}\textbf{55.5}
  & \cellcolor{blue!6}6.94
  & \cellcolor{blue!6}8.8 \\
  & Role-Shuffled RRB & $46.7{\pm}2.5$ & 6.98 & 8.7 \\
\bottomrule
\end{tabular}
}
\caption{Validation of block-relative RRB allocation on LLaDA-MoE. Math500 uses math-verify accuracy and HumanEval uses postprocessed pass@1. Role-Shuffled RRB reports mean accuracy and standard deviation over three allocation seeds; AvgK and Pairs are seed averages. It preserves the RRB expert-count distribution while randomly reassigning counts among processed token positions.}
\label{tab:role_shuffle}
\end{table}

As shown in Table~\ref{tab:role_shuffle}, Role-Shuffled RRB retains nearly the same AvgK as RRB Only but lowers Math500 and HumanEval accuracy by 3.7 and 8.8 points, respectively. The reported standard deviations are smaller than the corresponding mean gaps to RRB Only. Because the expert-count multiset is preserved within every forward, the comparison cannot be explained by the per-forward expert-count distribution alone. The remaining pair-count differences arise from NFE changes along the altered generation trajectories. Preserving the amount and distribution of expert computation is therefore insufficient when its correspondence with block-relative structural role is removed. This intervention supports the structural assignment used by RRB rather than its average budget alone.

\noindent\textbf{Ablation of active-block reallocation signals.}

We isolate the FGER ranking signal while holding the RRB profile, active-block expert-count levels, and group proportions fixed. The Frontier-Progress Score (FPS) uses $\score_i^t=F_i^tP_i^t$, where $F_i^t$ measures commitment proximity and $P_i^t$ measures limited confidence progress. Frontier Only and Progress Only retain one factor while removing the other. Confidence Only ranks tokens by lagged confidence $\conf_i^{t-1}$, whereas Uncertainty Only uses $1-\conf_i^{t-1}$. Since these configurations share the same available budget and reallocation levels, their comparison isolates how the active-block ranking signal places computation rather than how much computation is allocated.

\begin{table}[t]
\centering
{
\small
\setlength{\tabcolsep}{1.8mm}
\begin{tabular}{ccccc}
\toprule
\textbf{Benchmark} & \textbf{Signal} & \textbf{Acc} & \textbf{AvgK} & \textbf{Pairs (M)} \\
\midrule
\multirow{5}{*}{\textbf{Math500}}
  & Confidence Only & 40.0 & 6.3 & 21.6 \\
  & Uncertainty Only & 38.4 & 6.3 & 21.4 \\
  & Frontier Only & 39.2 & 6.3 & 21.7 \\
  & Progress Only & 40.0 & 6.3 & 21.5 \\
  & \cellcolor{blue!6}\textbf{FPS}
  & \cellcolor{blue!6}\textbf{41.8}
  & \cellcolor{blue!6}6.3
  & \cellcolor{blue!6}21.5 \\
\midrule
\multirow{5}{*}{\textbf{HumanEval}}
  & Confidence Only & 51.2 & 6.9 & 8.9 \\
  & Uncertainty Only & 51.8 & 6.9 & 8.8 \\
  & Frontier Only & 52.4 & 6.9 & 8.8 \\
  & Progress Only & 51.8 & 6.9 & 8.7 \\
  & \cellcolor{blue!6}\textbf{FPS}
  & \cellcolor{blue!6}\textbf{54.3}
  & \cellcolor{blue!6}6.9
  & \cellcolor{blue!6}9.0 \\
\bottomrule
\end{tabular}
}
\caption{Ablation of active-block reallocation signals on LLaDA-MoE. Math500 uses math-verify accuracy and HumanEval uses postprocessed pass@1. Configurations differ only in the reallocation signal; all other settings match the corresponding REFLEX experiments in the main paper.}
\label{tab:signal_ablation}
\end{table}

As shown in Table~\ref{tab:signal_ablation}, AvgK and selected pairs vary little across signals, so the quality differences primarily reflect where the shared active-block budget is assigned. FPS reaches 41.8 on Math500, exceeding the strongest single-factor result by 1.8 points, and reaches 54.3 on HumanEval, 1.9 points above the closest alternative. Frontier Only captures commitment proximity but cannot distinguish tokens making limited progress from those already resolving. Progress Only identifies slow confidence change but can prioritize tokens whose commitment is not imminent. Confidence Only and Uncertainty Only use the current confidence level without the temporal evidence contained in confidence progress. The multiplicative FPS assigns higher priority when commitment proximity coincides with limited progress. These matched-computation results support the joint ranking criterion used by FGER without treating FPS as a calibrated estimate of marginal refinement utility.

Taken together, the role-shuffling intervention and matched-computation signal ablation isolate the two levels of REFLEX. The former supports assigning the structural expert-count profile by block-relative role, while the latter supports using joint frontier and progress evidence to refine active-block allocation within that profile. Their effects therefore correspond to the structural prior and dynamic refinement evidence in the coarse-to-fine hierarchy.

\section{Hyper-parameter Sensitivity}

We vary one REFLEX hyper-parameter at a time on LLaDA-MoE while keeping the other three at their defaults. The default values are $\wcf=0.30$, $\vref=0.01$, $\tau=0.05$, and $q=0.25$. Each default column reuses the corresponding main experiment, and every other column changes only the indicated parameter. We report postprocessed HumanEval pass@1 and Math500 math-verify accuracy together with AvgK and selected expert-token pairs per sample. This setup separates changes in allocation amount from changes in how FGER ranks tokens within the active block.

\noindent\textbf{Commitment-frontier width $\wcf$.}
The commitment-frontier width determines the confidence interval below the commitment threshold that receives a high frontier factor. It also bounds the confidence change used in the one-forward forecast. Increasing $\wcf$ therefore broadens the region treated as close to commitment, but does not change the RRB profile or FGER group sizes.

\begin{table}[t]
\centering
\small
\setlength{\tabcolsep}{3pt}
\begin{tabular}{lccccc}
\toprule
\textbf{Metric} & \textbf{0.20} & \textbf{0.25} & \textbf{0.30} & \textbf{0.35} & \textbf{0.40} \\
\midrule
HE Acc.   & 51.83 & 51.83 & 54.27 & \textbf{57.32} & 54.88 \\
HE AvgK   & 6.93  & 6.95  & 6.95  & 6.95  & 6.95 \\
HE Pairs  & 8.73  & 8.85  & 8.95  & 8.96  & 8.66 \\
\midrule
M500 Acc.  & 40.60 & 40.60 & \textbf{41.80} & 41.00 & 40.20 \\
M500 AvgK  & 6.27  & 6.29  & 6.28  & 6.29  & 6.28 \\
M500 Pairs & 21.61 & 21.65 & 21.48 & 21.47 & 21.34 \\
\bottomrule
\end{tabular}
\caption{Sensitivity to the commitment-frontier width $\wcf$ on LLaDA-MoE. The default is $0.30$. HE denotes HumanEval postprocessed pass@1, and M500 denotes Math500 math-verify accuracy. Acc. is reported in percent and Pairs in millions; best task accuracies are shown in bold.}
\label{tab:hparam_wcf}
\end{table}

Table~\ref{tab:hparam_wcf} shows that AvgK remains within $0.02$ across the sweep for both tasks, while selected pairs also vary within a narrow range. The width therefore primarily changes which active-block tokens receive the higher count rather than the total allocation. Quality varies across tasks and settings, and we retain one shared default without task-specific selection.

\noindent\textbf{Velocity reference $\vref$.}
The velocity reference determines when recent confidence progress is considered limited. Raising $\vref$ makes the progress factor less selective with respect to positive confidence change, while lowering it emphasizes tokens whose confidence is nearly stationary or decreasing. Because FGER still assigns fixed proportions to its priority groups, this parameter changes their ordering criterion rather than their total budget.

\begin{table}[t]
\centering
\small
\setlength{\tabcolsep}{3pt}
\begin{tabular}{lccccc}
\toprule
\textbf{Metric} & \textbf{0} & \textbf{0.005} & \textbf{0.01} & \textbf{0.025} & \textbf{0.05} \\
\midrule
HE Acc.   & 53.05 & 52.44 & 54.27 & \textbf{56.71} & 54.27 \\
HE AvgK   & 6.94  & 6.94  & 6.95  & 6.94  & 6.94 \\
HE Pairs  & 8.81  & 8.75  & 8.95  & 8.96  & 8.94 \\
\midrule
M500 Acc.  & \textbf{42.00} & 40.20 & 41.80 & 41.00 & 40.60 \\
M500 AvgK  & 6.28  & 6.29  & 6.28  & 6.28  & 6.27 \\
M500 Pairs & 21.51 & 21.43 & 21.48 & 21.53 & 21.58 \\
\bottomrule
\end{tabular}
\caption{Sensitivity to the velocity reference $\vref$ on LLaDA-MoE. The default is $0.01$. HE denotes HumanEval postprocessed pass@1, and M500 denotes Math500 math-verify accuracy. Acc. is reported in percent and Pairs in millions; best task accuracies are shown in bold.}
\label{tab:hparam_vref}
\end{table}

As reported in Table~\ref{tab:hparam_vref}, AvgK changes by at most $0.01$ and pair counts remain close across all five values. Quality varies by task because the velocity reference changes which confidence trajectories are treated as slow. We retain the shared default across tasks without task-specific selection.

\noindent\textbf{Gate temperature $\tau$.}
The temperature controls the smoothness of both factors in FPS. Smaller values produce sharper changes around the commitment-frontier and velocity references, whereas larger values soften these boundaries and yield a more gradual ranking signal. The temperature does not alter the expert-count levels or the size of the reallocation groups.

\begin{table}[t]
\centering
\small
\setlength{\tabcolsep}{3pt}
\begin{tabular}{lccccc}
\toprule
\textbf{Metric} & \textbf{0.01} & \textbf{0.025} & \textbf{0.05} & \textbf{0.10} & \textbf{0.20} \\
\midrule
HE Acc.   & 52.44 & 51.22 & 54.27 & 53.05 & \textbf{56.71} \\
HE AvgK   & 6.94  & 6.94  & 6.95  & 6.94  & 6.94 \\
HE Pairs  & 8.84  & 8.96  & 8.95  & 8.93  & 8.81 \\
\midrule
M500 Acc.  & 39.60 & 38.80 & \textbf{41.80} & 39.00 & 40.00 \\
M500 AvgK  & 6.29  & 6.29  & 6.28  & 6.30  & 6.28 \\
M500 Pairs & 21.51 & 21.51 & 21.48 & 21.51 & 21.49 \\
\bottomrule
\end{tabular}
\caption{Sensitivity to the gate temperature $\tau$ on LLaDA-MoE. The default is $0.05$. HE denotes HumanEval postprocessed pass@1, and M500 denotes Math500 math-verify accuracy. Acc. is reported in percent and Pairs in millions; best task accuracies are shown in bold.}
\label{tab:hparam_tau}
\end{table}

Table~\ref{tab:hparam_tau} again shows nearly unchanged AvgK and selected pairs, indicating that $\tau$ mainly changes active-block ranking. Quality varies by task as gate smoothness changes, while the allocation amount remains stable. We retain the shared default without task-specific selection.

\noindent\textbf{Outer-group fraction $q$.}
The outer-group fraction sets the proportion of active-block tokens assigned to each of the high- and low-priority groups. Increasing $q$ applies the positive and negative expert-count offsets to more tokens. Since the two groups have equal size, their changes cancel and the active-block total remains fixed; $q$ controls the coverage of reallocation rather than its net budget.

\begin{table}[t]
\centering
\small
\setlength{\tabcolsep}{5pt}
\begin{tabular}{lccc}
\toprule
\textbf{Metric} & \textbf{0.125} & \textbf{0.25} & \textbf{0.375} \\
\midrule
HE Acc.   & 50.61 & \textbf{54.27} & 52.44 \\
HE AvgK   & 6.94  & 6.95  & 6.94 \\
HE Pairs  & 8.99  & 8.95  & 8.68 \\
\midrule
M500 Acc.  & 39.80 & \textbf{41.80} & 39.20 \\
M500 AvgK  & 6.28  & 6.28  & 6.27 \\
M500 Pairs & 21.54 & 21.48 & 21.69 \\
\bottomrule
\end{tabular}
\caption{Sensitivity to the outer-group fraction $q$ on LLaDA-MoE. The default is $0.25$. HE denotes HumanEval postprocessed pass@1, and M500 denotes Math500 math-verify accuracy. Acc. is reported in percent and Pairs in millions; best task accuracies are shown in bold.}
\label{tab:hparam_q}
\end{table}

As shown in Table~\ref{tab:hparam_q}, AvgK remains essentially fixed as expected from the symmetric reallocation. The default $q=0.25$ gives the highest quality on both evaluated tasks, while narrower and broader group coverage yield lower quality under nearly unchanged AvgK. The selected-pair differences are small and reflect the resulting trajectory changes rather than a direct change in the per-forward active-block total.

Across all four analyses, the allocation amount remains comparatively stable because $\wcf$, $\vref$, and $\tau$ modify the ranking signal, while $q$ preserves the active-block total through symmetric groups. Generation quality is more task dependent because each parameter changes which token trajectories receive additional computation. We therefore use one shared configuration across HumanEval and Math500 without task-specific hyper-parameter selection.

\end{document}